\documentclass{article}

\usepackage{amsmath,amssymb,amsthm}
\usepackage{microtype}
\usepackage{graphicx}
\usepackage{subfigure}
\usepackage{enumitem}
\usepackage{booktabs} 
\usepackage{xspace}
\usepackage{url}

\usepackage{hyperref}

\usepackage[accepted]{icml2021}

\icmltitlerunning{\alg: Deep Mixture Models for Long-term AQ Forecasting in the Delhi National Capital Region}

\newcommand{\alg}{{DELFI}\xspace}
\newcommand{\nef}{\text{NEF}\xspace}
\graphicspath{{figures/}}

\begin{document}

\twocolumn[
\icmltitle{\alg: Deep Mixture Models for Long-term Air Quality Forecasting in the Delhi National Capital Region}

\begin{icmlauthorlist}
\icmlauthor{Naishadh Parmar}{iitk}
\icmlauthor{Raunak Shah}{iitk}
\icmlauthor{Tushar Goswamy}{iitk}
\icmlauthor{Vatsalya Tandon}{iitk}
\icmlauthor{Ravi Sahu}{iitk}
\icmlauthor{Ronak Sutaria}{resp}
\icmlauthor{Purushottam Kar}{iitk}
\icmlauthor{Sachchida Nand Tripathi}{iitk}
\end{icmlauthorlist}

\icmlaffiliation{iitk}{Indian Institute of Technology, Kanpur, India}
\icmlaffiliation{resp}{Respirer Living Sciences, Mumbai, India}

\icmlkeywords{Machine Learning, Air Quality Forecasting, ICML}

\vskip 0.3in
]

\printAffiliationsAndNotice{}

\begin{abstract}
The identification and control of human factors in climate change is a rapidly growing concern and robust, real-time air-quality monitoring and forecasting plays a critical role in allowing effective policy formulation and implementation.
This paper presents \alg, a novel deep learning-based mixture model to make effective long-term predictions of Particulate Matter (PM) 2.5 concentrations. A key novelty in \alg is its multi-scale approach to the forecasting problem. The observation that point predictions are more suitable in the short-term and probabilistic predictions in the long-term allows accurate predictions to be made as much as 24 hours in advance. \alg incorporates meteorological data as well as pollutant-based features to ensure a robust model that is divided into two parts: (i) a stack of three Long Short-Term Memory (LSTM) networks that perform differential modelling of the same window of past data, and (ii) a fully-connected layer enabling attention to each of the components. Experimental evaluation based on deployment of 13 stations in the Delhi National Capital Region (Delhi-NCR) in India establishes that \alg offers far superior predictions especially in the long-term as compared to even non-parametric baselines. The Delhi-NCR recorded the 3rd highest PM levels amongst 39 mega-cities across the world during 2011-2015 and \alg's performance establishes it as a potential tool for effective long-term forecasting of PM levels to enable public health management and environment protection. 
\end{abstract}
\section{Introduction}
\label{sec:intro}
The global challenge of climate change demands a multi-faceted response. Rapid, robust, and real-time identification of human sources of climate change such as combustion, mining, and other activities is a key aspect in enabling agile and adaptive policy and regulatory decisions. Climate change positively correlates with air pollution levels, with air pollutants such as black carbon that constitute particulate matter (PM), methane, tropospheric ozone, and aerosols, also affecting the amount of incoming sunlight and contributing to global temperature rise and glacial degradation~\cite{Manisalidis2020}. For instance, fossil fuel and biomass combustion are the biggest source of black carbon aerosols that contribute to both PM levels as well as accelerating glacier melting in the Himalayas~\cite{Patella2018}.

In particular, PM2.5 refers to particulate matter with a diameter less than 2.5$\mu$m and includes combustion by-products, metals and organic materials. PM2.5 is classified as an atmospheric pollutant of high concern since owing to its small size and comparatively larger surface area, it can remain suspended for extended periods, be easily transported and infiltrate the pulmonary and circulatory systems, if inhaled. Chronic exposure to high PM2.5 levels has been linked to, heart attacks and strokes \cite{PopeIII2002}, and respiratory diseases such as lung cancer \cite{Kampa2008}. Maternal exposure to high PM2.5 levels elevates the risk of congenital heart defects in infants~\cite{Zhang2016}. Effective monitoring and regulatory control of PM2.5 levels presents the need for a robust long-term forecasting model for PM2.5, especially in high pollution regions such as the Delhi National Capital Region (NCR) in India where extremely high PM2.5 levels (484 on a scale on 500) led to the local government triggering a state of public health emergency on November 1, 2019 \cite{reuters2017}.



\textbf{Technical Contributions and Impact:} The primary technical contribution of this paper is \alg, a novel deep learning-based mixture model to perform long-term predictions of PM2.5 levels with  key technical contributions:
\begin{enumerate}[nolistsep]
    \item Adopting a novel novel multi-scale forecasting strategy employing probabilistic predictions for forecasts with horizons longer than 6 hours
    \item A light-weight technique employing pre-computed \nef features (see Sec~\ref{sec:method}) that allow spatio-temporal effects to be incorporated without additional expense
    \item A mixture model architecture with 3 components each comprised of a stack of LSTM networks with attention and an alternating optimization-based training strategy
    \item Forecasts for time horizons as large as 24 hours into the future that offer significantly improved accuracy compared to baseline methods while offering short-term (1-4 hour) predictions within sensor error levels
\end{enumerate}
\alg makes point short-term predictions (for up to 6 hours in the future) and probabilistic long-term predictions (for up to 48 hours in the future). Experiments indicate that the model can be used to reliably design long-term air quality forecasting and early-warning systems that enable rapid regulatory action of preventative nature. Given the close link between air-quality and the drivers of climate change, this can not only safeguard the health of citizens against the harmful effects of air pollution, but also mitigate the adverse impact of human activities on climate change.


\textbf{Related Works and Contributions of \alg in Context}: At a high level, most existing works use a monolithic Long Short Term Memory (LSTM)-based model to make short-term predictions, often next-hour but up to 6 hours. In contrast, \alg offers accurate probabilistic predictions upto 24 hours in advance that make it suitable for use in designing early warning systems. \citet{delhilstm} uses data on CO, NO2, NO, Ozone, PM2.5 and SO2 levels from high-grade monitors to perform next hour prediction of the same levels using LSTMs. In contrast, \alg uses PM1, PM10 and PM2.5 concentrations from low-cost sensors in addition to meteorological features to offer predictions over much longer time scales as well as MAE values that are 35\% lower. \cite{Chaudhary2018TimeSB} use LSTMs to make next-hour predictions but use additional sources of seasonal data such as holidays and traffic data to which \alg does not assume access. \citep{Kumar2011PCR} apply principal component regression to make short-term predictions at a single air quality monitoring station. In contrast, \alg ingests data from, and makes simultaneous predictions for, more than a dozen locations. Ensemble models have been also considered such as \cite{Liu2019} which incorporates 5 models but only offer predictions up to 6 hours in advance, and \cite{Bai2019} that uses an ensemble LSTM network to make next-hour predictions. In contrast, \alg makes use of fewer but diverse components in its mixture, and does a careful assignment of data used to train each component, finally using an aggregator network to assign attention weights to each component.



\section{\alg: \underline{D}eep Mixtur\underline{E}-models for \underline{L}ong term air-quality \underline{F}orecast\underline{I}ng}
\label{sec:method}

\textbf{Data}: \alg is trained on air pollution as well as meteorological data from 13 stations in the region of Delhi-NCR, India, latitudes 28° 27' 15" to 28° 38' 40" and longitudes 77° 4' 26" to 77° 19' 40". Data collection was done at intervals of 1 hour in the time period of 1 November 2018 00:00:00 to 28 March 2019 23:00:00. PM2.5 concentrations that appeared extremely elevated were not removed as outliers from the data so as to enable the model to be trained and tested on predicting spikes in PM2.5 levels.

\textbf{Training Features}: \alg uses a total of 9 features, each available as a time-series at each station, to perform forecasting:  (i) PM1, (ii) PM10, (iii) PM2.5, (iv) Temperature, (v) Humidity, (vi) Visibility, (vii) Wind speed, (viii) Wind direction and (ix) the Net external flow (NEF). Of these, the first 8 features are standard and available from the monitoring stations or else standard APIs. However, the last NEF feature (described below) was engineered to allow \alg to take spatio-temporal effects of long-range air flow into account. A \emph{data-point} $w_{i, t}$ is defined as a six-hour window (ending at timestamp $t$) of these pollutant and meteorological features taken at one-hour intervals at station $i$. The true PM2.5 concentration value at station $i$ at timestamp $t$ will be denoted by $PM_{i, t}$ below.

\textbf{The \nef Feature}: This feature attempts to capture in a relatively inexpensive manner, the effect of PM2.5 concentrations at other stations given wind direction and velocities. For a station $a$ and time $t$, this feature is defined as
\[
\nef_{a, t} = \frac{1}{1+e^{-x}}, x = \sum_i PM_{i, t} \times V_{i, t}\times cos(\theta_a^i - \phi_{i,t}),
\]
where $PM_{i, t}$, $V_{i,t}$ and $\phi_{i, t}$ are respectively the true PM2.5 concentration, wind speed and wind bearing at station $i$ at time $t$ and $\theta_a^i$ is the bearing between station $a$ and station $i$ based on the World Geodetic System (WGS84).

\textbf{A Key to Long-term Forecasting}: As discussed in Section~\ref{sec:intro}, most existing work attempts to make short term forecasts (up-to 6 hours). Long term forecasting rapidly deteriorates in quality possibly due to the lack of complete knowledge of all factors affecting air quality as well as the chaotic nature of aerodynamic systems. However, \alg makes an observation that when making longer-term forecasts, say 24 hours in advance, a \textit{point} prediction becomes less critical. Specifically, if making a 24 hour forecast at 1100hrs today, it is not critical to know the PM2.5 levels at exactly 1100hrs tomorrow. Rather, the \textit{distribution} of PM2.5 levels around 1100hrs (e.g. in the period 0900-1300 hrs, are the levels likely to remain mild or can they spike) becomes more important both from the point of policy formulation as well as modulating personal behavior. To exploit this observation, \alg uses air-quality categorization set forth by regulatory authorities in India, namely 0-30 (Good), 30-60 (Satisfactory), 60-90 (Moderately polluted), 90-120 (Poor), 120-250 (Very poor) and 250+ (Severe) to create 6 \textit{bins}. Thus, when making long-term forecasts, say in the above example, \alg predicts a discrete probability distribution over these 6 bins that indicate the probability of hourly PM2.5 values within the 0900-1300 hrs time window falling in those bins. Experiments find that probabilistic predictions perform better for long-term forecasts.

\textbf{Data Setup}: To make point predictions, that are more useful in the short term, \alg trains on the \textit{residuals} as is common in literature~\cite{zheng2015forecasting} i.e. rather than learning to predict $PM_{i, (t+1)}$ directly for a next-hour point prediction, \alg learns to predict $\Delta PM_{i, \left(t+1\right)} = PM_{i, \left(t+1\right)} - PM_{i, t}$ instead. Thus, for next-hour point predictions, the set $\{w_{i, t}, \Delta PM_{i, \left(t+1\right)}\}$ defines the training set. To make longer term point predictions, say given a $s$-hour \textit{horizon} (i.e. predicting $s$ hours into the future), \alg simply makes next-hour predictions $s-1$ times, sliding the 6-hour window by a single hour each time using its own predicted values. This allows \alg to use a single model to make point predictions at various horizons say 1-hour, 2 hours, etc and not learn a separate model for every horizon.
However, in line with past works, our experiments show that point predictions become severely inaccurate beyond 6 hour horizons. \alg adopts probabilistic predictions for longer-term horizons say 8, 12, 24 and 48 hours, as discussed above. For such a long-term horizon $s$ and data-point $w_{i, t}$, a window of length $s$ centred at $t+s$ i.e. $PM_{i, \left(t+j\right)}, \forall j \in [t+s/2, t+3s/2)$ is used to create a normalized histogram denoted by $h_{i,t+s}$ of these values binned into the 6 bins described above. The set $\{w_{i, t}, h_{i,t+s}\}$ then defines the training set. A different training set is thus created for every value of horizon $s$. For point as well as probabilistic predictions, data points created out of the first 85\% timestamps of each station's data were used as training data and those created out of the suffix 15\% of each station's data were used as test data.



\begin{figure}[t]
\centering
\includegraphics[width=0.8\linewidth]{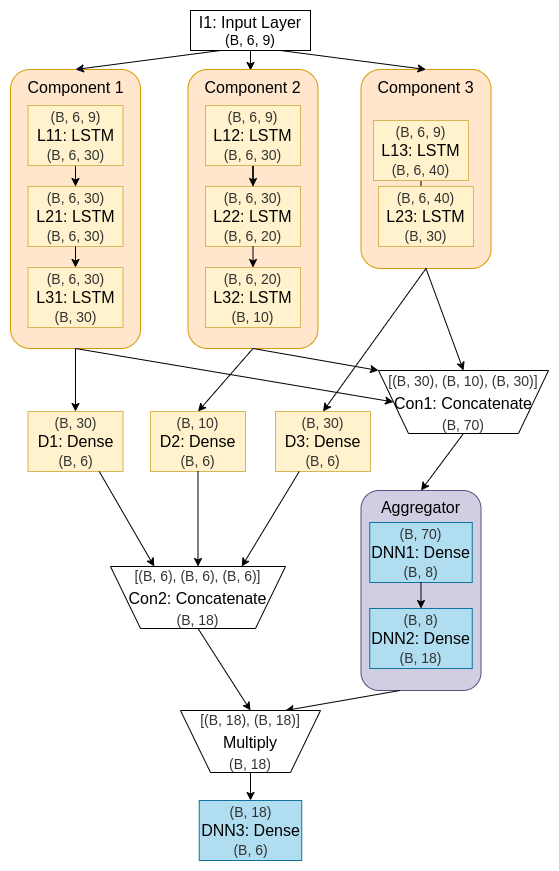}
\caption{Model architecture used by \alg for long-term probabilistic predictions. $B$ is batch-size. The blue and orange portions of the network are trained alternately as described in Algorithm~\ref{algo:train}.}
\label{fig:amtold-long}
\end{figure}

\textbf{Model and architecture}: Initial experiments were conducted with 13 separate LSTM-based model being trained on data from each of the 13 stations in the deployment. However, this strategy neither took advantage of the much larger amount of overall data, nor did models learnt for one station do well in predicting values for another station. However, given the diversity of the stations (some were prominent PM2.5 hotspot candidates whereas others reported much milder PM2.5 values), it was also expected that a single model may struggle to address these extremes. \alg's solution is mixture model that consults 3 components, each being a stack of LSTM networks with attention weights being learnt for each component (see Figure~\ref{fig:amtold-long}). More specifically, each component consists of a series of stacked LSTMs which take in features $w_{i,t}$ as sequential input and offers a sequential output. \alg develops two distinct mixture models: one for short-term predictions and one for long-term predictions. A fully connected layer is used as the \textit{aggregator} that offers the attention weights. For short-term predictions (see Figure \ref{fig:amtold-short} in the appendix), each component predicts a certain residual value, for example $\hat{\Delta PM}_{i,(t+1)}$ for next-hour predictions and weights are assigned to the output of each component. For long-term predictions, the output is used to scale the concatenated sequences and passed through another fully connected layer to get the final histogram output. These large-capacity architectures allow the 13 diverse stations to adaptively choose a component best suited to make predictions for its data.



\begin{table*}[t]
\centering
\begin{tabular}{| c || c | c | c | c | c | c | c | c | c |}
\hline
 & 1 hr & 2 hr & 3 hr & 4 hr & 5 hr & 6 hr & 8 hr & 12 hr & 24 hr \\
\hline
\hline
KNN & 9.27 & 16.42 & 23.20 & 30.55 & 38.72 & 47.86 & 68.53 & 120.01 & 326.30\\
\hline
Linear & 10.01 & 23.19 & 39.08 & 63.99 & 103.19 & 162.71 & 366.81 & 1356.74 & 37569.30\\
\hline
\alg & 10.03 & 17.49 & 25.28 & 35.22 & 47.90 & 62.58 & 95.07 & 175.43 & 450.91\\
\hline
\end{tabular}
\caption{Mean absolute error (MAE) in $\mu\textrm{g}$ $\textrm{m}^{-3}$ between point predictions and actual PM2.5 values for all methods and various horizon values. For short term point-predictions (e.g. 1-4 hour horizons) KNN, linear model as well as \alg are competitive and offer acceptable performance. In particular, \alg MAE values for $s = 1$hr are within sensor error levels. Performance deteriorates to unacceptable levels for all methods for long- term horizons e.g. 8+ hrs which indicates that point predictions are ill-suited for long-term forecasting.}
\label{tab:short-term-predictions}
\end{table*}


\begin{algorithm}[t]
	\caption{\alg training via Alternating Optimization}
	\label{algo:train}
	\begin{algorithmic}[1]
	    \STATE Perform pre-training (stations grouped into 3 categories using variance of residual and each group assigned to a component)
		\FOR{$t$ in $n_{epochs}$}
		\STATE Freeze aggregator and subsequent dense layer
		\STATE Train mixture model components on data-points assigned to it for $n_t$ iterations 
		\STATE Freeze mixture model components 
		\STATE Train aggregator and subsequent dense layer for $m_t$ iterations 
		\ENDFOR
	\end{algorithmic}
\end{algorithm}

\textbf{Pre-training}: The variance in the residual values $\Delta PM_{i, \left(t+1\right)}$ was computed for all stations. Using this statistic, stations were grouped into three categories. Each group was assigned one component in the mixture and data from within the group was used to pre-train model parameters for that particular component. This division was used to decide the data points used to pre-train the three individual components before starting the main training step. As a result, stations prone to spiking PM2.5 values were likely to cluster together into one group, with other stations with more gentle variations in PM2.5 levels falling into another.

\textbf{Training}: An alternating optimization procedure reminiscent of the EM algorithm was adopted to train both the fully connected layer (aggregator) and individual components of the mixture described in Algorithm~\ref{algo:train}. The mixture model components and aggregator have been colored differently (Figures~\ref{fig:amtold-long} and \ref{fig:amtold-short}) to highlight the alternating nature of the training algorithm. For point predictions, the mean squared loss between the predicted residual $\hat{\Delta PM}_{i, \left(t+1\right)}$ and the actual residual $\Delta PM_{i, \left(t+1\right)}$ was used to train the model. For probabilistic long-term predictions, the Kullback-Liebler divergence loss between the predicted and actual histograms was used to train the model. The Adam optimizer~\cite{Kingma14} was used to learn the parameters of the model with a learning rate of $0.005$. The training loss was found to reasonably converge within $n_{epochs} = 10$ and $n_t = m_t = 10$ iterations during experiments.


\begin{table}[t]
\begin{tabular}{| c || c | c | c | c | c |}
\hline
 Time (hrs) & 6$\pm$3 & 8$\pm$4 & 12$\pm$6 & 24$\pm$12 & 48$\pm$24\\
\hline
\hline
KNN & 1.33 & 1.80 & 2.40 & 1.40 & 1.17\\
\hline
\alg & \textbf{0.32} & \textbf{0.89} & \textbf{1.38} & \textbf{0.59} & \textbf{0.53}\\
\hline
\end{tabular}
\caption{KL divergence between predicted and actual histograms for various horizon lengths. Linear models were unable to provide meaningful predictions and were excluded from comparison. \alg offers KL divergence values that are at least 42\% and upto 76\% smaller than those offered by the KNN algorithm.}
\label{tab:long-term-predictions}
\end{table}

\section{Experimental Results}
\label{sec:exps}
To validate \alg's performance, baseline strategies popular in air-quality calibration and forecasting literature \cite{amt2021} such as parametric (linear) and non-parametric (k-nearest neighbors or KNN) models were considered. These baselines often offer acceptable performance for real-time calibration as well as short-term forecasting but rapidly deteriorate on the more challenging task of long-term forecasting. The results offered by linear and KNN models as well as \alg on the point prediction task for various horizon lengths are presented in Table \ref{tab:short-term-predictions}. Table \ref{tab:long-term-predictions} presents results on the task of making probabilistic forecasts. The KNN algorithm was suitably modified to output probabilistic predictions by averaging ouputs in the identified neighborhood. The results for point predictions are presented in terms of the mean absolute error (MAE) between the actual PM2.5 concentration and the predicted PM2.5 concentration, whereas results for probabilistic predictions are presented in terms of the Kullback-Liebler divergence (KL divergence) between the actual and predicted histograms. In both cases, lower values correspond to better performance.

Table \ref{tab:short-term-predictions} demonstrates that for short horizon values (1-4 hours), \alg offers performance that is comparable to the non-parametric method KNN. However, for long horizon values (8+ hours), point predictions from all models are extremely inaccurate with those from the linear model being especially poor. Although KNN does offer marginally better performance than \alg for larger horizons, the outputs of neither algorithm can be deemed acceptable. Rather than a statement on the algorithm, Table \ref{tab:short-term-predictions} shows that point predictions are ill-suited for long-term forecasting. However, the trends are vastly different in Table \ref{tab:short-term-predictions} where \alg offers much superior performance as compared to the KNN algorithm, with as much as 76\% less KL divergence values.

\section{Discussion and Impact on Climate Change}
This paper presents \alg, a novel algorithm that introduces several technical innovations such as the use of probabilistic predictions for long-term forecasting and using a mixture model trained using an alternating strategy. Intuitively, a low bias model is preferred for short-term predictions to encode local features properly that explains KNN's good performance on small horizons (1-4 hrs). However, for long-term predictions, as unpredictability of the system grows, a low-variance method is preferable instead that KNN does not offer. \alg seems to offer a suitable balance between bias and variance allowing it to perform well in both regimes. Experimental results suggest that \alg offers predictions reliable enough to design early-warning systems based on long-term forecasts. In highly polluted regions of the globe such as the Delhi-NCR, such systems offer citizens a chance to modulate their own personal behavior e.g. avoiding outdoor activities, but can also enable regulatory authorities to take immediate preventative action as well as effect long-term policy shift. Given the close link between air-quality and the drivers of climate change discussed in Section~\ref{sec:intro}, this can not only safeguard the health of citizens but also mitigate the adverse impact of human activities on climate change in the longer term.

\bibliographystyle{plainnat}
\bibliography{delfi}
\clearpage
\appendix

\section{Appendix}

\begin{figure}[h]
\centering
\includegraphics[width=\linewidth]{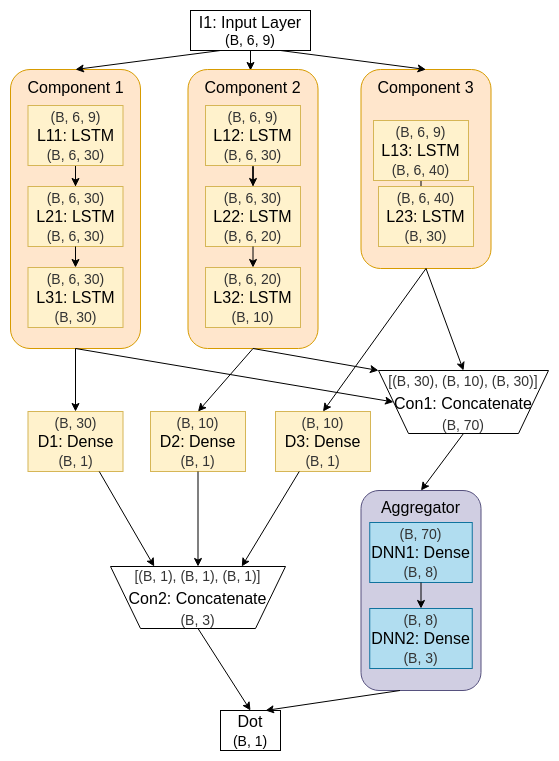}
\label{fig:amtold-short}
\caption{Model architecture used by \alg for short-term point predictions. $B$ is batch-size. The blue and orange portions of the network are trained alternately as described in Algorithm~\ref{algo:train}.}
\end{figure}
\end{document}